\documentclass{article}

\usepackage{spconf}
\usepackage{amsfonts}   
\usepackage{amssymb}  
\usepackage{amsmath}  
\usepackage{amsthm}
\usepackage{color}
\usepackage{graphicx}
\usepackage{url}  
\usepackage{cite}
\usepackage{enumitem}
\usepackage{soul}
\usepackage{stmaryrd}
\usepackage{bm}
\usepackage{multirow}
\usepackage[aboveskip=2pt]{subcaption}
\usepackage{lipsum}
\usepackage{hyperref}

\input{mysymbol.sty}
\newtheorem{problem}{Problem}

\ninept

\title{Power allocation for wireless federated learning \\ using graph neural networks}
\name{
Boning Li$^\star$, 
Ananthram Swami$^\dag$, 
and Santiago Segarra$^\star$
\thanks{
Research was sponsored by the Army Research Office and was accomplished under Cooperative Agreement Number W911NF-19-2-0269. 
The views and conclusions contained in this document are those of the authors and should not be interpreted as representing the official policies, either expressed or implied, of the Army Research Office or the U.S. Government. 
The U.S. Government is authorized to reproduce and distribute reprints for Government purposes notwithstanding any copyright notation herein.\newline
E-mails: \{boning.li, segarra\}@rice.edu, ananthram.swami.civ@army.mil.}
}
\address{$^\star$Rice University, USA \hspace{1cm} $^\dag$US Army DEVCOM Army Research Lab., USA}

\begin{document}
\setlength{\abovedisplayskip}{3pt}
\setlength{\belowdisplayskip}{3pt}

\maketitle

\begin{abstract}
\vspace{-.5em}
We propose a data-driven approach for power allocation in the context of federated learning (FL) over interference-limited wireless networks. 
The power policy is designed to maximize the transmitted information during the FL process under communication constraints, with the ultimate objective of improving the accuracy and efficiency of the global FL model being trained.
The proposed power allocation policy is parameterized using a graph convolutional network and the associated constrained optimization problem is solved through a primal-dual algorithm.
Numerical experiments show that the proposed method outperforms three baseline methods in both transmission success rate and FL global performance.
\end{abstract}


\vspace{-1em}
\section{Introduction}\label{s:intro}
\vspace{-.5em}

Federated learning (FL) is a collaborative machine learning methodology in which a model is trained based on multiple local datasets without explicitly exchanging data samples~\cite{mcmahan2017communication}.
This makes it particularly relevant in settings where privacy concerns are important as in autonomous driving~\cite{li2021privacy}, digital healthcare~\cite{xu2021federated}, or policy constraints on data sharing in a coalition military network. 
The FL pipeline starts with the server broadcasting a global model to selected workers who update that model using their own local data and, after one or more {\it local} training  iterations, transmit the updated model back to the server. 
Upon receiving all local models, the server aggregates them into a new global model, which completes one {\it global} iteration (or simply called an FL iteration). 
A typical FL process may involve several global iterations until the global loss converges, thus, the quality of the communication between the server and the individual workers is a fundamental aspect in determining the success of FL.

Due to the growing computational power of mobile devices and the increasing speed and reliability of wireless connections, there is a fast-growing interest in studying FL scenarios where the communications are through wireless  networks~\cite{tran2019federated,wang2019adaptive,dong2020communication,girgis2021shuffled,hanna2021quantization}. 
This opens a wide range of novel problems where one seeks to optimally manage the wireless network in response to the specific demands of FL.
For instance, one can try to perform power allocation -- which is essential to reduce multiuser interference and ensure the quality of wireless services~\cite{chiang2008power,choi2016power,matthiesen2020globally} -- with the objective of accelerating the convergence of the global model being trained.
Indeed, there has been a recent trend of analyzing wireless problems that arise in this FL setting.
For example,~\cite{chen2020joint} analyzes the impact of the wireless channel on FL convergence rate, based on which a joint optimization scheme of power, worker, and resource management is proposed for static channels. 
Without prior knowledge of the wireless channels, multi-armed bandit worker scheduling has been studied to reduce FL time consumption in the presence of heterogeneous data~\cite{xia2020multi} and fluctuating resources~\cite{yoshida2020mab}.
A common limitation in these works is the assumption of orthogonal channels without multiuser interference, which simplifies the problem of optimal power control but reduces the practical applicability of the derived solutions.

Over the past few years, machine learning has thrived on solving many problems in wireless communications, including resource allocation with various objectives (e.g., to maximize system utility in data rate, energy efficiency, or proportional fairness)~\cite{amiri2018machine,zhang2021scalable,nasir2019multi,kumar2021adaptive}.
In particular, since graph neural networks (GNNs) are designed to facilitate learning from graph data, they are well-suited for wireless networks~\cite{zhao2021distributed,eisen2020optimal,shen2019graph,chowdhury2021unfolding,chowdhury2021ml}.
For example, to maximize sum-rate in interference networks, a suboptimal power allocation algorithm was enhanced with augmented parameters learned by graph convolutional networks (GCNs) for both single-input single-output~\cite{chowdhury2021unfolding} and multiple-input multiple-output~\cite{chowdhury2021ml} systems. 
The empirical success of GCNs inspires us to leverage their learning capabilities to develop a power allocation policy that is efficient and tailored to the specific requirements of FL. 
In this paper, we devise (to the best of our knowledge) the first GCN-based power allocation policy for FL over (interference-limited) wireless networks.

\vspace*{1mm}
\noindent\textbf{Contributions.}
The contributions of our paper are twofold:\\
1)~We state a formal problem for power allocation under FL requirements and propose a solution based on a GCN parameterization and a primal-dual iterative approach, and \\
2)~Through numerical experiments on simulated wireless channels and a real-world machine learning dataset we validate the effectiveness of the proposed FL power allocation approach.

\vspace{-1em}
\section{System Model and Problem Formulation}\label{s:pre}
\vspace{-.5em}

We formally introduce FL and our wireless network model in Sections~\ref{ss:fl} and~\ref{ss:wl}.
In Section~\ref{ss:prob}, we provide a precise formulation of power allocation for FL as an optimization problem.

\vspace{-1em}
\subsection{Federated learning over wireless networks}\label{ss:fl}
\vspace{-.5em}

Our FL system consists of a server and $L$ mobile worker devices, where the server and workers communicate through wireless links.
Each worker $i$ collects local data $\bbX_i=[\bbx_i^{(1)},...,\bbx_i^{(k_i)}]$ associated with labels $\bby_i=[y_i^{(1)},...,y_i^{(k_i)}]$. 
The total number of training data samples is $K=\sum^{L}_{i=1}k_i$ and the data $\ccalX=\{\bbX_1,...,\bbX_L\}$ are assumed to be independent and identically distributed (i.i.d.).
Each worker is learning a parametric local function $\Phi(\cdot; \bbw_i)$ that seeks to map $\bbX_i$ to $\bby_i$, where $\bbw_i$ denotes the trainable parameters specific to worker $i$.
Each worker aims to minimize a local loss $f(\bbX_i,\bby_i;\bbw_i):=\sum_{j=1}^{k_i} \ell(\hat{y}_i^{(j)},y_i^{(j)})$, where $\hat{y}_i^{(j)} = \Phi(\bbx_i^{(j)}; \bbw_i)$ and $\ell$ is some loss function specific to the label space such as quadratic loss for continuous labels and cross-entropy for classification.
Overall, the FL process seeks to solve the following optimization problem~\cite{chen2020joint},\looseness=-1 
\begin{equation}\label{e:p1}
    \begin{aligned}
        \min_{\bbw_1,...,\bbw_L} & \,\,\sum\limits^{L}_{i=1} f(\bbX_i,\bby_i; \bbw_i),\\
        \text{s.t.} \quad & \bbw_1=...=\bbw_L=\bbw_{\mathrm{gl}},\\
    \end{aligned}
\end{equation}
where $\bbw_{\mathrm{gl}}$ is the global FL model aggregated by the server. 
More precisely, in an iterative two-step process: i) each worker updates their parameters $\bbw_i$ based on local data and uploads these to the server and ii) the server aggregates the local models into a global one via $\bbw_{\mathrm{gl}}=\frac{1}{K}\sum_{i=1}^L k_i \bbw_i$, which is then broadcast to every worker for them to update their local copies.
This last step guarantees that the constraint in~\eqref{e:p1} is satisfied.

The wireless setting adds additional complexities to the more classical FL setup.
Since workers upload local models over wireless links that could be unreliable in realistic environments, the transmission may be corrupted. 
Erroneous transmissions may have a negative impact on FL performance because errors will be aggregated into the global model and contaminate all workers in the next iteration. 
Additionally, delays in the wireless transmissions from some of the workers can significantly slow down global convergence.

\vspace{-1em}
\subsection{Multiuser interference network}\label{ss:wl}
\vspace{-.5em}

Consider the uplink of a network with $L$ single-antenna mobile workers and an $n_R$-antenna base station (BS).
Denoting the signal transmitted by worker $j$ as $s_j \in \mbC$, the received signal at the BS is ${\bbr} = \sum^L_{j=1}{\bbh}_{j} s_j + {\bbz}$, where ${\bbh}_{j} \in \mbC^{n_R}$ is the channel from $j$ to the BS and ${\bbz} \sim \ccalN_\ccalC(0, \sigma_i^2)$ is the additive complex Gaussian noise at the receiver. 
Throughout the paper, we assume perfect channel information.
For a given vector of assigned powers ${\bbp}=[p_1,...,p_L]$ and assuming the application of match filters at the receiver, the signal-to-interference-plus-noise ratio (SINR) of the link from worker $i$ to the BS is given by
\begin{equation}\label{e:sinr}
    \SINR_i = \frac{\alpha_i p_i}{1+\sum_{j\neq i}\beta_{i,j} p_j}, \quad\forall i=1,...,L,
\end{equation} 
where the channel gain and interference coefficients are computed as
\begin{equation}
\alpha_i=\frac{\|{\bbh}_{i}\|^2}{\sigma_i^2} \quad \text{ and} \quad \beta_{i,j}=\frac{|{\bbh}^H_{i} {\bbh}_{j}|^2}{\sigma_i^2\|{\bbh}_{i}\|^2}. \nonumber
\end{equation} 
We define the channel-state information (CSI) matrix  ${\bbH}\in \mbR^{L\times L}$ with $\alpha_i$ as its diagonal entries and $\beta_{i,j}$ as off-diagonal entries, i.e., $H_{i,i}=\alpha_i$ and $H_{i,j}=\beta_{i,j}$ for $i\neq j$. 
Notice that by introducing the CSI matrix, the system can now be interpreted as a directed graph with workers as its nodes and $\bbH$ as its adjacency matrix. 

The achievable data rate for worker $i$ with bandwidth $B$ is $R_i(\bbp,\bbH) = B\log(1+\SINR_i)$, and its transmission delay is 
\begin{equation}\label{e:tau}
    \tau_i = \frac{Z(\bbw_i)}{R_i(\bbp,\bbH)},
\end{equation}
where $Z(\bbw_i)$ is the size of the transmitted local model $\bbw_i$ in bits.
We assume single-packet transmissions and define the packet error rate 
\begin{equation}\label{e:per}
    \PER_i(\bbp,\bbH)=1-e^{-\frac{m}{\SINR_i}},
\end{equation}
with $m$ being a waterfall threshold~\cite{chen2020joint,xi2011general}.
To simplify notation, we often denote the error probability as $q_i=\PER_i(\bbp,\bbH)$.

Cyclic redundancy check (CRC) is leveraged every time a packet containing $\bbw_i$ arrives at the BS and, with probability $\tdq_i=1-q_i$, the transmission is error-free. 
We define a system-level performance metric $g(\cdot)$ as the weighted sum of the probabilities of successful transmissions
\begin{equation}\label{e:wsq}
    g(\tbbq) = \sum_{i=1}^L \omega_i \tdq_i,
\end{equation}
with $\omega_i$ being a weight assigned to worker $i$ accounting for the data quantity (e.g., we set $\omega_i=k_i/K$ in our simulations) or other factors such as worker preference or reliability.

\vspace{-1em}
\subsection{Problem formulation}\label{ss:prob}
\vspace{-.5em}

Our goal is to find an instantaneous power allocation policy $p:\mbR^{L\times L}\rightarrow \mbR^L$ such that the allocated powers $\bbp=p(\bbH)$ for varying channels $\bbH$ achieve the best possible global FL model as stated in~\eqref{e:p1} while satisfying the constraints imposed by our physical system as explained in Section~\ref{ss:wl}.

More precisely, notice that under the CRC mechanism, only the correctly transmitted local models will be aggregated by the server.
Hence, the update of the global model at the server will be a function of the channel and the power allocation as
\begin{equation}
\bbw_{\mathrm{gl}}(\bbp, \bbH) = \frac{\sum_i k_i \bbw_i S_i(\bbp,\bbH) }{ \sum_i k_i S_i(\bbp,\bbH)},
\end{equation}
where $S_i(\bbp,\bbH)$ is the binary outcome of the transmission by worker $i$.
In other words, $S_i(\bbp,\bbH)=1$ if $p_i>0$ and the corresponding transmission is successful, and $S_i(\bbp,\bbH)=0$ otherwise.

Under the i.i.d. assumption, the global model will benefit from aggregating the largest number possible of local models $\bbw_i$. 
Hence, we adopt $g(\tbbq)$ in~\eqref{e:wsq} as a proxy for the quality of the global FL model.
We can now formally state our main problem.

\begin{problem}\label{P:main}
Determine the optimal power allocation policy $p^*: \mbR^{L\times L}\rightarrow \mbR^L$ that solves the following optimization problem
\begin{subequations}
    \begin{alignat*}{3}
        p^* = & \argmax_{p}\quad  g\left( \EH[\mathbf{1}-\PER(p(\bbH),\bbH)] \right), \label{e:p2}\tag{P1}\\
        \text{s.t.} \quad  & r_{0,i} \leq \EH[R_i(p(\bbH),\bbH) \, | \, R_i(p(\bbH),\bbH) > 0], \forall i, \label{e:p2:a}\tag{P1-a}\\
         & p(\bbH) \in [0,P_{\max}], \,\,\, \forall \,\, \bbH, \label{e:p2:b}\tag{P1-b}
    \end{alignat*}
\end{subequations}
where $\PER(p(\bbH),\bbH) = [\PER_1(p(\bbH),\bbH), \ldots, \PER_L(p(\bbH),\bbH)]^\top$, $\ccalH$ is some channel distribution of interest, and $r_{0,i}$ is a prespecified minimum level of desired expected data rate for user $i$.
\end{problem}

To better understand \eqref{e:p2} notice that we are trying to maximize $g(\cdot)$ applied to the expected probabilities of successful transmissions given a distribution $\ccalH$ from which the channels are drawn. 
We seek to maximize this objective subject to two natural constraints.
In~\eqref{e:p2:a} we require the data rate at each worker that is actually transmitting to exceed (in expectation) a prespecified rate $r_{0,i}$. 
In light of expression~\eqref{e:tau}, this requirement can be reinterpreted as demanding a delay smaller than a maximum allowable transmission delay for every transmitter with non-zero allocated power.
Moreover, in~\eqref{e:p2:b} we guarantee that the instantaneous allocated power $p(\bbH)$ for every worker does not exceed an upper bound $P_{\mathrm{max}}$.

It should be observed that optimally solving~\eqref{e:p2} is particularly challenging.
Apart from having a non-convex objective [cf.~\eqref{e:sinr} and~\eqref{e:per}] there are at least two additional sources of complexity:
i)~We are optimizing over the space of power allocation functions, which is inherently infinite-dimensional, and ii)~We are not given a specific channel or sets of channels but rather we have to consider expected values over a whole distribution of channels.
We explain how our proposed solution addresses these challenges in the next section.

\vspace{-1em}
\section{Primal-Dual Graph Convolutional Power Network}\label{s:alg}
\vspace{-1em}

Our method (termed \underline{p}rimal-\underline{d}ual \underline{g}raph convolutional power \underline{net}work, or PDGNet) addresses the two aforementioned challenges by parameterizing the space of power allocation functions through graph neural networks (Section~\ref{ss:usca}) and by relying on a primal-dual restatement of the problem (Section~\ref{ss:pd}).

\vspace{-1em}
\subsection{Power policies using graph neural networks}\label{ss:usca}
\vspace{-.5em}

In order to render the maximization in~\eqref{e:p2} tractable, we parameterize a subset of all possible policies $p$ using a GCN and instead maximize over the parameters of this neural architecture.
To be more precise, we define $p(\bbH)=\Psi(\bbH;\bbTheta)$ where $\Psi$ is a $T$-layer GCN with trainable weights $\bbTheta$.
The input to $\Psi(\bbH;\bbTheta)$ is $\bbZ^{(0)} = P_{\mathrm{max}} \mathbf{1}$ and an intermediate $t$th layer of the GCN is given by
\begin{equation}\label{e:gcn}
	\bbZ^{(t)} = \sigma_t\left(
	\bbD^{-\frac{1}{2}}\bbH\bbD^{-\frac{1}{2}}\bbZ^{(t-1)}\bm{\Theta}^{(t)}\right).
\end{equation}
In~\eqref{e:gcn}, $\bbD = \diag(\bbH \mathbf{1})$ is the degree matrix,
${\bbTheta}^{(t)} \in \mathbb{R}^{d_{t-1} \times d_{t}}$ are the trainable parameters,
$d_{t-1}$ and $d_{t}$ are respectively the dimensions of the output features of layers $t-1$ and $t$, and $\sigma_t(.)$ is an activation function. 
The specifics of the number of layers $T$, the hidden dimensions $d_t$, and the activation functions $\sigma_t$ used in our experiments are further detailed in Section~\ref{s:exp}.

In principle, one can choose many parameterizations of the power policy functional space.
Our particular choice in~\eqref{e:gcn}, not only achieves good performance in practice but is also driven by the following three appealing characteristics.

\vspace{1mm}
\noindent
\emph{Permutation equivariance.} 
GCNs are composed of graph filters, which are similar to convolutional filters in classical convolutional neural networks~\cite{segarra2017optimal}.
These graph filters learn to perform appropriate local aggregations through trainable parameters.
Importantly, our parameterization entails a permutation equivariant power allocation policy~\cite{gama2020stability}.
Putting it differently, the allocated power is not affected by the indexing of our workers, which is a key feature since this indexing is arbitrary.

\vspace{1mm}
\noindent
\emph{Flexibility to varying number of workers.} 
Dynamic wireless environments and on-request termination of FL participation create possible variations in the communication network size during FL.
This requires the power allocation policy to be able to accommodate a number of workers $L'$ different from the $L$ workers with which it was trained.
This distinguishing capability is achieved by relying on GCNs [cf.~\eqref{e:gcn}] and is numerically showcased in Section~\ref{s:exp}.

\vspace{1mm}
\noindent
\emph{Distributed deployment.} 
The GCN-based power policy is naturally amenable to a distributed deployment~\cite{segarra2017optimal}.
However, it should be noted that centralized training is necessary.
Upon training completion, the whole set of GCN parameters ${\bm \Theta}$ must be broadcast to every worker.
Although the offline training phase has to be centralized, a decentralized online phase can facilitate its scalability to systems with large number of users.

\vspace{-1em}
\subsection{Primal-dual augmented constrained learning}\label{ss:pd}
\vspace{-.5em}

Having introduced the GCN parameterization of $p$, we restate~\eqref{e:p2} in a way that is amenable to a primal-dual solution
\begin{subequations}
    \begin{alignat*}{3}
        &\max_{\bm{\Theta},\tbbq,\bbr}\quad  g(\tbbq), \label{e:p3}\tag{P2}\\
        &\text{s.t.} \quad  \tbbq \leq \EH[\mathbf{1}-\PER(\Psi(\bbH;\bbTheta),\bbH)], \label{e:p3:a}\tag{P2-a}\\
         & r_i \leq \EH[R_i(\Psi(\bbH;\bbTheta),\bbH) \, | \, R_i(\Psi(\bbH;\bbTheta),\bbH) \!> \!0], \label{e:p3:b}\tag{P2-b}\\
         & r_i \in [r_{0,i},+\infty), \,\, \forall i, \label{e:p3:c}\tag{P2-c}\\
         & \Psi(\bbH;\bbTheta) \in [0,P_{\max}], \,\,\, \text{for all} \,\, \bbH. \label{e:p3:d}\tag{P2-d}
    \end{alignat*}
\end{subequations}
We enforce~\eqref{e:p3:d} simply by ensuring that the image of the last non-linearity in our GCN is contained in $[0,P_{\max}]$.
To treat the remaining constraints, we formulate the Lagrangian dual problem by introducing nonnegative multipliers $\bblambda_q\in\mbR_{+}^{L}$ and $\bblambda_r\in\mbR_{+}^{L}$ associated with constraints~\eqref{e:p3:a} and~\eqref{e:p3:b}, respectively. 
To simplify the notation, let $\mbE_{\text{H}}[f_q]$ and $\mbE_{\text{H}}[f_r]$ denote the expectation terms in constraints~\eqref{e:p3:a} and~\eqref{e:p3:b}, respectively.
Moreover, we denote the feasible region in~\eqref{e:p3:c} as $ \ccalR := [\bbr_0,+\infty)$.
The Lagrangian of~\eqref{e:p3} is then given by
\begin{equation}
    \begin{aligned}
        \ccalL_{\Psi}(\bbTheta,\tbbq,&\bbr,\bblambda_q,\bblambda_r)=\\ 
        &g(\tbbq) + \bblambda_q^{\top} (\mbE_{\text{H}}[f_q]-\tbbq) + \bblambda_r^{\top} (\mbE_{\text{H}}[f_r]-\bbr).
    \end{aligned}
\end{equation}
Following standard primal-dual solutions of optimization problems~\cite{boyd2004convex}, this Lagrangian motivates the following iterative primal-dual learning approach
\begin{subequations}
    \begin{alignat}{3}
        \bbTheta_{k+1} &= \bbTheta_k + \gamma_{\Theta,k}(\nabla_{\Theta}\mbE_{\text{H}}[f_q]\bblambda_{q,k} + \nabla_{\Theta}\mbE_{\text{H}}[f_r]\bblambda_{r,k}), \label{e:lag:1}\\
        \tbbq_{k+1} &= \tbbq_{k} + \gamma_{q,k}(\nabla_{q}g(\tbbq)), \label{e:lag:2}\\
        \bbr_{k+1} &= \mathrm{proj}_{\ccalR}\left( \bbr_k + \gamma_{r,k} (-\bblambda_{r,k}) \right), \label{e:lag:3}\\
        \bblambda_{q,k+1} &= [\bblambda_{q,k} - \gamma_{\lambda_q,k} (\mbE_{\text{H}}[f_q] - \tbbq_{k+1}) ]_{+}, \label{e:lag:4}\\
        \bblambda_{r,k+1} &= [\bblambda_{r,k} - \gamma_{\lambda_r,k} (\mbE_{\text{H}}[f_r] - \bbr_{k+1}) ]_{+}, \label{e:lag:5}
    \end{alignat}
\end{subequations}
where the terms $\gamma_{*,k}$ denote update step sizes of the corresponding variables in the $k$th iteration.
In our implementation, we approximate the expected values in~\eqref{e:lag:1}, \eqref{e:lag:4}, and \eqref{e:lag:5} through their empirical counterparts by drawing several channels from $\ccalH$.

Leveraging the result in~\cite[Theorem 1]{eisen2019learning} one can show that under mild conditions (such as $\ccalH$ being a non-atomic distribution and the feasibility region having at least an interior point) the duality gap depends linearly on the approximation capabilities of the parameterization of our policy function.
Putting it differently, if our proposed $\Psi(\bbH;\bbTheta)$ is a good parameterization of the power allocation policy space, then the solution of the primal-dual iterations in~\eqref{e:lag:1}-\eqref{e:lag:5} will be close to the optimal solution of our original problem~\eqref{e:p2}.
From a practical perspective, in Section~\ref{s:exp} we compare PDGNet to a primal-dual solution based on a multi-layer perceptron (MLP) -- whose universal approximation capabilities are well established~\cite{hornik1989multilayer} -- and empirically show that PDGNet indeed outperforms graph-agnostic learning methods.

\vspace{-1em}
\section{Numerical experiments}\label{s:exp}
\vspace{-1em}

\begin{figure*}[t] 
\centering
    \begin{subfigure}{0.24\linewidth}
        \includegraphics[width=\linewidth]{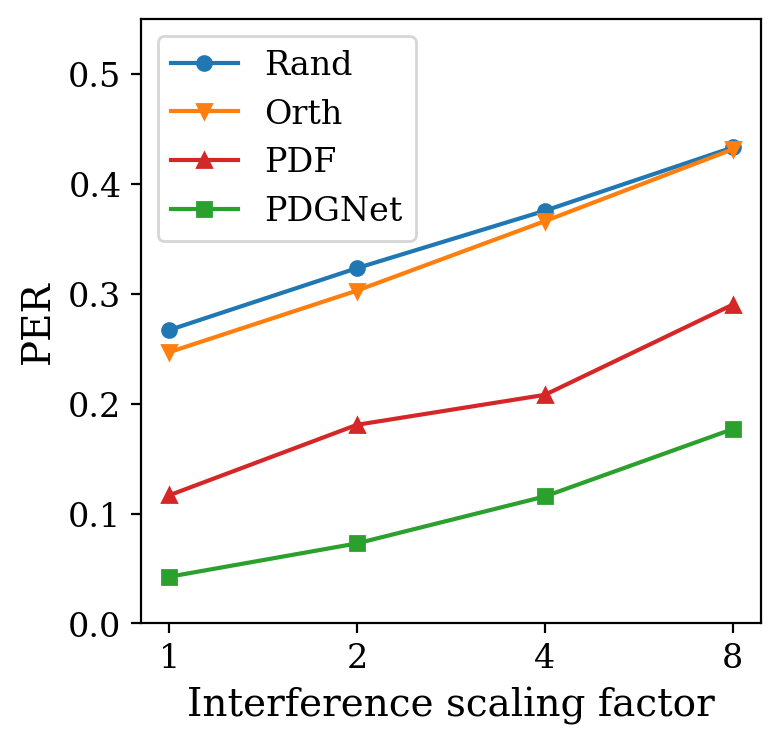}
        \caption{}
        \label{ff:1:r1}
    \end{subfigure}\hfil
    \begin{subfigure}{0.24\linewidth} 
        \includegraphics[width=\linewidth]{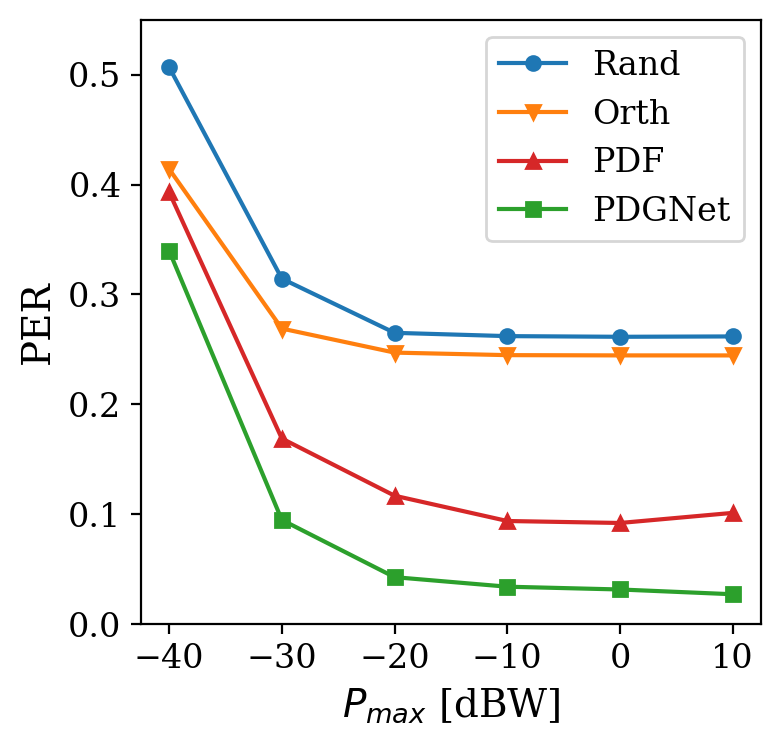}
        \vspace{-1em}
        \caption{}
        \label{ff:1:r2}
    \end{subfigure}\hfil
    \begin{subfigure}{0.24\linewidth} 
        \includegraphics[width=\linewidth]{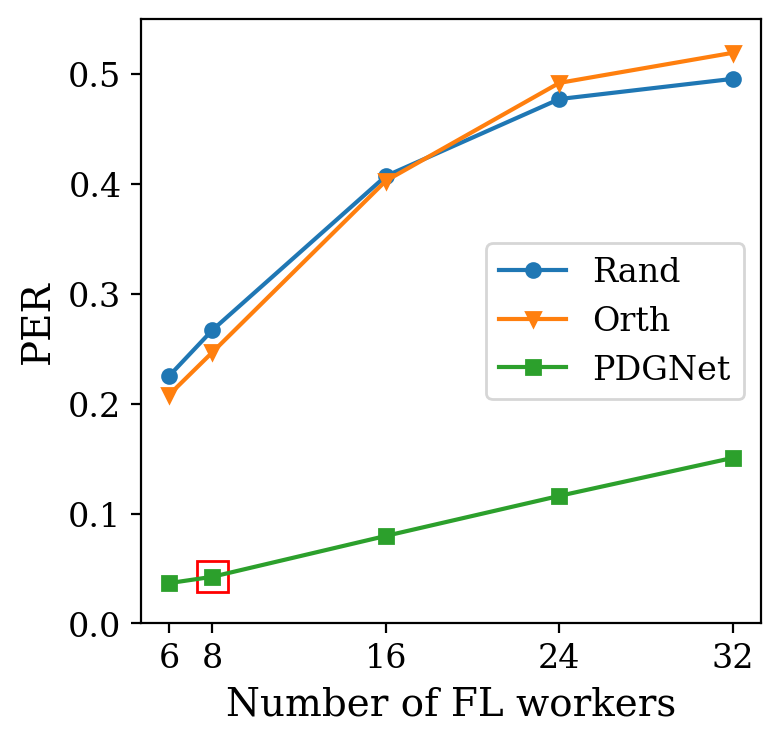}
        \caption{}
        \label{ff:1:r3}
    \end{subfigure}\hfil
    \begin{subfigure}{0.27\linewidth}
        \includegraphics[width=\linewidth]{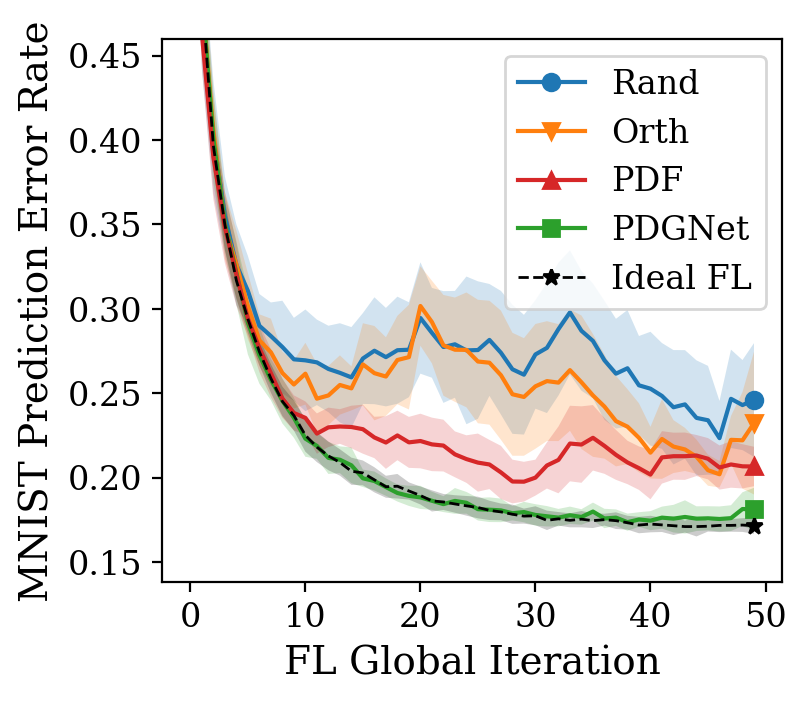}
        \caption{}
        \label{ff:1:r4}
    \end{subfigure}
\vspace{-1em}
\caption{
PERs and performance of the FL model achieved by different power allocation policies.
The first group of comparison is based on the weighted-sum PER of selected workers against
(a)~interference strength,  
(b)~maximum power constraint value, and
(c)~network size;
the last one (d) displays the MNIST classification error of the FL global model at the end of each FL iteration, including the ideal FL (assuming no transmission loss) performance.
The proposed PDGNet outperforms the baseline power allocation strategies in all the metrics considered.
}\label{f:1}
\vspace{-1em}
\end{figure*}

We evaluate the proposed power allocation method PDGNet\footnote{\scriptsize{Code to reproduce the reported results can be found at \url{https://github.com/bl166/WirelessFL-PDGNet}.}} based on two metrics, namely the wireless transmission performance and the prediction performance of the FL global model.
For wireless channels, we employ the same data generator as in~\cite{matthiesen2020globally} to simulate wideband spatial (WBS) interference networks\footnote{\scriptsize{\url{https://github.com/bmatthiesen/deep-EE-opt}}.} served by $M=1$ BS with $n_R=10$ antennas.
We assume negligible downlink communication cost while focusing on the uplink transmission from workers to servers.
A total of 2000 channel realizations are generated for training (to update model parameters) and validation (to select the best performing epoch for test) in equal splits, and another 1000 left out for testing.
We select different values for $P_{\max}$ ranging from -40 to 10$\dBW$ (default at -20$\dBW$ unless otherwise specified).
With a default number of $L=8$ workers, the power policy network in PDGNet is a 5-layer GCN with intermediate feature dimensions $\{16, 32, 64, 16, 2\}$. 
The activation functions $\sigma_t$ for intermediate layers are set as ELU while for the last layer we apply a scaled sigmoid to guarantee that the output power is contained in $[0,P_{\max}]$.
Learning rates $\gamma_{\Theta}=10^{-3}$ and $\gamma_{\{q,r,\lambda_q,\lambda_r\}}=10^{-4}$ are fixed for all iterations.
A number of 1000 primal-dual learning epochs are performed in an unsupervised manner.
We compare our method to two model-based and one learning-based baselines:\\
\textbf{Rand} is a random power policy that assigns uniformly random power $\sim\ccalU(0,P_{\max})$ to every worker.\\
\textbf{Orth} is the optimal power control for orthogonal channels, resulting in all workers transmitting at maximum power. \\
\textbf{PDF} is a multi-layer perceptron (MLP) power policy model trained following the same primal-dual algorithm as PDGNet. 
It takes as input the flattened $\bbH$ with $P_{\max}$ appended, propagates it through 5 hidden layers with \{128, 256, 64, 16, 8\} neurons and ELU activations in between.
The final non-linearity is the same as that of PDGNet.

In Rand and Orth, after power is determined for all candidate workers, a subset of workers satisfying the constraints are selected to participate in the current FL iteration.
In the learning-based methods PDF and PDGNet, participating workers are not explicitly selected;
instead, workers assigned zero transmit power will automatically drop. 
In this way, our experiments confirm that the constraints are satisfied on unseen channels through appropriate training.

For the FL task, we consider the multiclass classification of the MNIST dataset~\cite{lecun1998gradient} by a single-layer (50 hidden nodes) feedforward neural network with hyperbolic tangent activation and cross entropy loss. 
The number of data samples at each worker is drawn from a uniform distribution $\ccalU(20,200)$. 
The batch size for local training is 16, and the optimizer is Adam with learning rate $10^{-3}$.
Within each global iteration, workers performs a single local training epoch with their own on-device data. 
A dynamic wireless environment is simulated by generating a new channel realization from the same WBS distribution at the beginning of each FL iteration.  
We report the average performance on a 1000-sample test set over 5 random runs.

\noindent\textbf{Transmission over different interference magnitude.}
We scale the interference coefficients in the original network by factors of \{1,2,4,8\} and compare the transmission performance of candidate methods in those scenarios (Figure~\ref{ff:1:r1}).
The plotted metric is the weighted-sum PER of all the workers that transmit. 
Although Orth is optimal for orthogonal channels under transmission delays~\cite{chen2020joint}, here in  nonorthogonal channels its performance is comparable with that of Rand.
Both primal-dual learning methods transmit with smaller error rate, while PDGNet can transmit with even fewer errors than the topology-agnostic PDF in all tested interference intensities. 

\noindent\textbf{Transmission performance versus maximum power values.}
When $P_{\max}$ is small enough, all power policies tend to allow maximum transmit power since the multiuser interference is small as well, thus their performance is close. 
When $P_{\max}$ grows and approaches approximately $-10 \dBW$, interference becomes the major limiting factor.
As a result, we observe in Figure~\ref{ff:1:r2} that the advantage of PDGNet's policy is more conspicuous for larger $P_{\max}$ settings.

\noindent\textbf{Transmission for different number of workers.}
To demonstrate that the proposed method is robust to changes in network size, we test the transmission performance of the candidate models in different sized FL systems.
Note that PDF is not eligible for this experiment because its MLP power policy only works with fixed input and output dimensions. 
In contrast, PDGNet is suitable due to the versatility of the GCN. 
In Figure~\ref{ff:1:r3}, PDGNet constantly guarantees better wireless uplink transmission than the other two baselines in \{6,8,16,24,32\}-worker systems, though trained only on the 8-worker system (marked with a red box).

\noindent\textbf{Prediction performance of the global FL model.}
Finally, we run the (trained) power control methods iteratively within the FL pipeline. 
At the beginning of every FL iteration, a new wireless channel is drawn. 
Each power policy is then applied to the new channels to determine the transmit power of all workers, following which the PER is computed for each transmission. 
Based on those PERs, we randomly draw whether each transmission was successful or not.
If a transmission is deemed successful, then the corresponding local model will take part in the global weighted aggregation during the current FL iteration. 
The multiclass prediction errors of a left-out test set is computed based on the aggregated global models. 
This result is presented in Figure~\ref{ff:1:r4}, showing that FL based on PDGNet power control outperforms all competitors in both efficiency (fastest speed of convergence) and accuracy (smallest final gap compared to the ideal FL).

\vspace*{-1em}
\section{Conclusions and Future Work}\label{s:end}
\vspace*{-.5em}

We developed a data-driven approach for FL power control based on GCNs and demonstrated the efficacy of the proposed method in (approximately) solving the associated non-convex constrained optimization problem.
Moreover, we have highlighted additional appealing characteristics of the proposed approach such as robustness to the indexing of the workers and being able to accommodate changes in the number of workers even after training.
Future research avenues include: 
(i)~Enforcing additional constraints such as bounds on the expected energy consumption and 
(ii)~Exploring more challenging application scenarios such as large-scale communication networks, imperfect channel information, and non-i.i.d. data at the workers. 

\bibliographystyle{IEEEbib}
{
    \bibliography{IEEEabrv,references}
}
\end{document}